 \documentclass[pmlr,twocolumn,10pt]{jmlr} 

\usepackage{booktabs}
\usepackage[load-configurations=version-1]{siunitx} 

\theorembodyfont{\upshape}
\theoremheaderfont{\scshape}
\theorempostheader{:}
\theoremsep{\newline}

\jmlryear{2021}
\jmlrworkshop{Machine Learning for Health (ML4H) 2021} 

\title[RadBERT-CL]{RadBERT-CL: Factually-Aware Contrastive Learning For Radiology Report Classification}

  \author{%
   \Name{Ajay Jaiswal, The University of Texas at Austin} \Email{ajayjaiswal@utexas.edu}\\
   \Name{Liyan Tang, The University of Texas at Austin} \Email{lytang@utexas.edu}\\
   \Name{Meheli Ghosh, Central University of Gujarat} \Email{mehelighosh69@gmail.com}\\
   \Name{Justin F Rousseau, Dell Medical School} \Email{justin.rousseau@austin.utexas.edu}\\
   \Name{Yifan Peng, Weill Cornell Medicine} \Email{yip4002@med.cornell.edu}\\
   \Name{Ying Ding, The University of Texas at Austin} \Email{ying.ding@ischool.utexas.edu}
  }

\begin{document}

\maketitle

\begin{abstract}
Radiology reports are unstructured and contain the imaging findings and corresponding diagnoses transcribed by radiologists which include clinical facts and negated and/or uncertain statements. Extracting pathologic findings and diagnoses from radiology reports is important for quality control, population health, and monitoring of disease progress. Existing works, primarily rely either on rule-based systems or transformer-based pre-trained model fine-tuning, but could not take the factual and uncertain information into consideration, and therefore generate false positive outputs. In this work, we introduce three sedulous augmentation techniques which retain factual and critical information while generating augmentations for contrastive learning. We introduce RadBERT-CL, which fuses these information into BlueBert via a self-supervised contrastive loss. Our experiments on MIMIC-CXR show superior performance of RadBERT-CL on fine-tuning for multi-class, multi-label report classification. We illustrate that when few labeled data are available, RadBERT-CL outperforms conventional SOTA transformers (BERT/BlueBert) by significantly larger margins (6-11\%). We also show that the representations learned by RadBERT-CL can capture critical medical information in the latent space.
\end{abstract}

\section{Introduction}
\label{sec:intro}

Chest radiography is a critical medical imaging technique used for diagnosis, screening, and treatment of many perilous diseases. Radiology reports are documented by radiologists after examining a patient's medical history and diagnostic imaging, and represent complex anatomical and medical terms written for healthcare providers, along with indications of the presence or absence of any disease. Classifying radiology reports according to their description of abnormal findings is important for quality assurance and can mitigate the risks of diagnostic radiation exposure in children [24]. Additionally, the Precision Medicine Initiative (PMI) initiated by NIH and multiple research centers has highlighted the importance of text mining techniques to enable cohort phenotyping of patients for population health  \citep{Shin2017ClassificationOR}. Classifying radiology reports can help to identify patient cohorts and enable precision medicine on a large scale. Labeling radiology reports with disease types can also assist in the development of deep learning applications for automated-diagnosis  \citep{rajpurkar2017chexnet,han2021using,yao2018learning}.

In recent works, rule-based systems have been developed to categorize radiology reports into disease categories using medical domain knowledge and careful feature engineering. ChestX-ray14  \citep{Wang_2017}, MIMIC-CXR  \citep{johnson2019mimiccxrjpg}, and OpenI  \citep{DemnerFushman2016PreparingAC} are some of the largest radiology datasets available, and many classification algorithms have been developed based on the training sets provided by these datasets to classify reports into diseases. CheXpert  \citep{irvin2019chexpert} is an automated rule-based labeler consisting of three stages: mention extraction, mention classification, and mention aggregation, to extract observations from the free text radiology reports to be used as structured labels for the images. CheXBert  \citep{smit2020chexbert} uses the labels extracted by CheXpert to fine-tune BERT transformer along with $\sim 1000$ manually annotated reports to classify radiology reports. While these methods have shown great advancements, they cannot capture many critical and factual information (especially negated statements). Negated statements in a radiology report can lead to false positive classifications and therefore should be treated with caution. Also negated statements provide rich information that should be captured and integrated into the classification algorithms.

\begin{table*}
    \centering
    \caption{Examples from the set of rules in our Info-Preservation Module for Negation and Uncertainty Detection and their corresponding matching sentences.}
    \begin{tabular}{p{\textwidth}}
    \toprule
    \textbf{BACKGROUND:} Radiographic examination of the \textcolor{red}{chest}. clinical history: 80 years of age, male.  PA AND LATERAL CHEST, \_\_\_\\\\
    \midrule
    \textbf{FINDINGS:} \textcolor{red}{Heart size} and \textcolor{red}{mediastinal contours} are normal.  The \textcolor{red}{right hilum} is asymmetrically enlarged compared to the \textcolor{red}{left hilum} but has a similar size and configuration compared to a baseline radiograph \_\_\_ \_\_\_.  A chest CT performed in \_\_\_ demonstrated \textcolor{red}{no evidence} of a \textcolor{red}{right hilum mass}, and the observed asymmetry is \textcolor{red}{probably} due to a combination of a slight rotation related to mild \textcolor{red}{scoliosis} and a prominent \textcolor{red}{pulmonary vascularity}.
        
    \textcolor{red}{Lungs} are slightly hyperexpanded but grossly \textcolor{red}{clear of} \textcolor{red}{pleural effusions}.\\
    \vspace{0.07cm}
    \textbf{IMPRESSION:}  \textcolor{red}{No} radiographic evidence of \textcolor{red}{pneumonia}.

    \end{tabular}
    \vspace{-0.5cm}
    \label{tab:example}
\end{table*}

        
        

 Motivated by the success of contrastive learning in computer vision  \citep{chen2020simple, he2020momentum, chen2020improved, grill2020bootstrap, robinson2020contrastive} to improve on the learning of feature representation in latent space, we propose to pre-train transformers using contrastive learning before the end-to-end fine-tuning for classification of radiology reports. Medical reports contain many critical and factual information such as the presence/absence of a disease (see Table \ref{tab:example} for more details). This information is central for making a classification decision, and many other downstream tasks such as Report Generation  \citep{zhang2020radiology}, Report Summarization  \citep{zhang2020optimizing}, etc. Most existing approaches do not handle uncertainty/negation information explicitly, and depend on the deep learning models to capture them. We identified that the SOTA transformers such as Bert \citep{devlin2019bert}, BlueBert \citep{peng2019transfer}, do not perform well at capturing uncertainity/negation information in latent space. Considering the significance of these critical information for both interpretability and performance improvement of deep learning models, we introduce RadBERT-CL, a pre-trained model using contrastive learning which can capture critical medical and factual nuances of radiology reports. It trains BlueBert \citep{peng2019transfer} with the radiology report dataset and captures its fine-grained properties, in order to improve performance of report classification task at the fine-tuning stage. We introduce three novel data augmentation techniques at the sentence and document level, which can retain the critical medical concepts and factual information present in radiology reports while generating positive and negative pairs for contrastive learning. 

\begin{figure*}[ht]

\caption{(a) Pre-training architecure of RadBERT-CL using contrastive learning. Two separate data augmentation views are generated using the augmentation techniques described in Section \ref{augmentation}. Both views (query and key) are passed through RadBERT-CL, which is a transformer-based encoder $f(.)$, and a projection head $g(.)$. RadBERT-CL is trained to maximize agreement between the two augmented views using contrastive loss. (b) Fine-tuning Model architecture of RadBERT-CL. The model consists of $14$ linear heads corresponding to 14 disease concepts. Among them, $13$ linear heads can predict 4 outputs, while linear head corresponding to ``No Finding'' can predict $2$ outputs.}

\includegraphics[width=\textwidth, trim=0em 6em 2em 2em ]{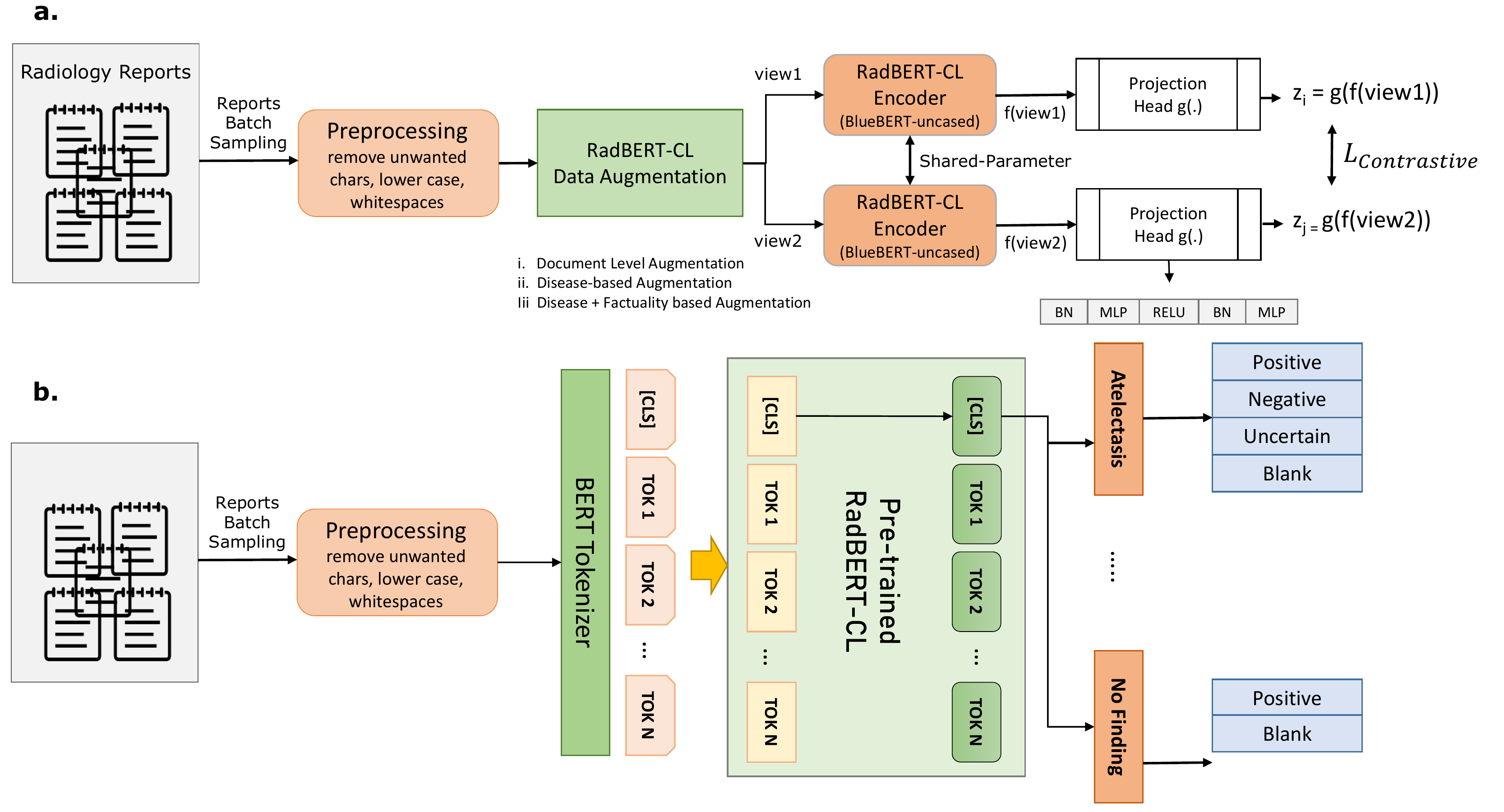}
\label{fig:structure}
\end{figure*}

RadBERT-CL outperforms the previous best reported CheXbert labeler  \citep{smit2020chexbert} with  0.5\% improvement on F1-score without any need for hight quality manual annotation during training (note that the baseline \citep{smit2020chexbert} has claimed their results very close to human-level performance). We evaluated our system using 687 expert-annotated reports, same as CheXbert \citep{smit2020chexbert}. We find that representations learned by RadBERT-CL are more informative, can capture and distinguish critical information present in the radiology reports. The improvements on F1-measure are more significant if few manually annotated data are available. This is particularly important since obtaining manually annotated data in medicine is extremely difficult and costly. In this case, our algorithm can achieve 6-11\% improvements on disease classification. The highlights of our contributions are:
\begin{itemize}

    \item We propose two novel data augmentation techniques which retain factual and critical medical concepts, identified by our semi-rule based Info-Preservation Module, while generating positive and negative keys for contrastive learning.
    \item We show that our model RadBERT-CL is able to learn and distinguish fine-grained medical concepts in latent space, which cannot be captured by SOTA pre-trained models like BERT, and BlueBert.
    \item We apply contrastive learning for radiology report classification task and show improvements on the state-of-the-art methods. We use weakly-labeled data during our training and evaluate our system using 687 high-quality reports manually labelled by radiologists.
    \item Lastly, we evaluate our model performance when a few data labels are available for training and show that our model outperforms significantly by 6-11\% improvements in disease classification task. 
\end{itemize}

\begin{table}[h]
\centering
\caption{Explanation of class value predicted by RadBERT-CL for disease observations}
\begin{tabular}{p{1.4cm}p{6cm}}
\toprule
\textbf{Blank} & observation not mentioned in the report\\
\midrule
\textbf{Positive} & observation mentioned and its presence is confirmed\\
& eg. definite focal consolidation is seen in lungs\\
\midrule
\textbf{Negation} & observation mentioned and its absence is confirmed\\
& eg. the lungs are clear of any focal consolidation\\
\midrule
\textbf{Uncertain} & observation mentioned with uncertainty\\
& eg. signs of parenchymal changes suggesting pneumonia\\
\bottomrule
\end{tabular}
\label{tab:train_params}
\end{table}

\section{Related Work}
\subsection{Contrastive Learning} Contrastive learning (CL) seeks to learn effective representations by maximizing the agreement between two augmentations from one example and minimizing the agreement of augmentations from different instances. CL has been recently explored in computer vision and graph Neural Network due to its success in self-supervised representation learning. However, CL still receives limited interest in the NLP domain. The main reason is the discrete nature of text and it is hard to define and construct effective positive pairs. Several works have explored ways to perform augmentations.  \citep{Fang2020} back-translated source sentences to create sentence-level positive augmentations, which maintain semantic meaning of the source sentence.  \citep{Wu2020} integrated four sentence-level augmentation techniques, namely word and span deletion, reordering and synonym substitution, to increase models' robustness. 

\begin{table*}
    \centering
     \caption{Examples highlighting the selection of positive and negative keys for a given anchor sentence using two different approaches for Sentence-Level Contrastive Learning. For Disease-based Augmentation, a given anchor sentence with disease concept d, any other sentence from any report mentioning d can be taken as positive example. In Disease + Factuality Based Augmentation, we incorporate mentions from our negation or uncertainty dictionary along with disease concept while generating augmentation pairs.}
    \begin{tabular}[ht]{l}
    
    \hline
    \textbf{a. Disease-based augmentation}\\
    \hline
    \textit{Anchor/Query} : definite \underline{focal consolidation} is seen in left side of lungs\\
    \textit{Positive Key} \ \ : there is a \underline{focal consolidation} at the left lung base adjacent to the lateral hemidiaphragm\\
    \textit{Negative Key} \ : there are low lung volumes and mild \underline{bibasilar atelectasis}\\
    
    \hline
    \textbf{b. Disease + Factuality based augmentation}\\
    \hline
    \textit{Anchor/Query} : definite \underline{focal consolidation} is seen in left side of lungs\\
    \textit{Positive Key}\ \ \  : there is a \underline{focal consolidation} at the left lung base adjacent to the lateral hemidiaphragm\\
    \textit{Negative Key} \ : the lungs are \textcolor{red}{clear of} any \underline{focal consolidation}\\
    \hline
    \end{tabular}

    \label{tab:augmentationexample}
\end{table*}

\subsection{Factual Correctness and Consistency} Factual correctness and factual consistency are key requirements for medical reports. Keeping factual information and avoiding hallucinations could support medical decision-making process. These requirements have been recently explored in NLP tasks, especially in abstractive text summarization. \citep{Zhang2020} directly took factual correctness as a training objective in their system via reinforcement learning. On the other hand,  \citep{Falke2019} and,  \citep{Goyal2020} used textual entailment to detect factual inconsistency based on the assumption that summary should be entailed by the source document.  \citep{zhu2021} built a knowledge graph containing all the facts in the text, and then fused it into the summarization process.

\section{Methods} 

\subsection{Problem Formulation}
Radiology report classification is a multi-class multi-label classification problem, which classifies radiology reports into different disease observations (e.g., cardiomegaly, effusion, mass, edema). Following  \citep{smit2020chexbert}, we label each report $r^k$ in MIMIC-CXR dataset with a 14-dim vector $y = [y_1,\ y_2,\ y_3,\ ...\ ,\ y_{14}]$ of observations, where observations $y_1 - y_{13}$ can take any value from the following 4 classes : blank, positive, negative, and uncertain. For $y_{14}$, which corresponds to \textit{No Finding} (no pathology is found in the scope of any of 13 disease observations), the classifier takes value from only 2 classes: blank, and uncertain.

\subsection{Data Augmentation}
\label{augmentation}

In computer vision, it has been verified that contrastive learning benefits from strong data augmentation techniques like random cropping, rotation, blurring, color distortion, etc  \citep{chen2020simple, tian2020contrastive, chen2020improved, he2020momentum}. However, in NLP, generating data augmentation is comparatively difficult due to the discrete representation of words, and it is unknown what kind of augmentation will benefit  noise-invariant representational learning.  \citep{fang2020cert} used back-translation to perform sentence augmentation while  \citep{wu2020clear} explored four different basic augmentation techniques: word and span deletion, reordering, and substitution. While these methods have shown improvements on some SentEval and GLUE benchmarks, they cannot be directly applied to generating augmentations for radiology reports. Radiology reports contain critical and factual information and that need to be preserved while generate augmentations. Table \ref{tab:example} presents an example of radiology report in which we have highlighted the information such as \textit{chest, left hilum, pulmonary vascularity, clear of, no evidence, pneumonia, etc.}

Through augmentation, it is likely that  \citep{wu2020clear} dropped critical words or phrases which can lead to a completely different diagnosis. For example, dropping negation words, such as \textit{No}, can lead to a diagnosis suggesting the presence of pleural effusion, and it can have negative consequences during our downstream task of disease classification. Also, as suggested by  \citep{fang2020cert}, back-translation cannot provide satisfactory results for the medical data because back-translation models have limited the cross-language translation ability for domain specific texts. 

In order to ensure that critical and factual information is preserved while generating augmentations, we define an \textit{Info-Preservation} module, which identifies and preserves facts during augmentation generation. We propose sentence-level and document-level augmentation techniques, to effectively pre-train our RadBERT-CL architecture.

\subsubsection{Info-Preservation Module}
Radiology reports consist of many important radiology concepts such as diseases, body parts, etc. In order to preserve them during augmentation, we develop a rule-based tool similar to Dynamic-LCS  \citep{raj2020solomon} to greedily match concepts in RadLex ontology  \citep{Langlotz2006RadLexAN} on sequences of the lemmatized tokens in the reports (longer matches are returned when possible). For capturing the presence of negation of any concept, we manually create a dictionary of $30$ negation indicator keywords such as: \textit{not, without, clear of, ruled out, free of, disappearance of, without evidence of, no evidence of, absent, miss}. Following  \citep{Chen_2018}, we create a dictionary of uncertainty keywords with a wide range of uncertain types, from speculations to inconsistencies present in the reports. We design a set of pattern matching rules following  \citep{Wang_2017} for identifying sentences containing negation or uncertainty. Appendix Table \ref{tab:ruleexample} presents some examples of our rules and the matched sentences from the radiology reports. While generating augmentations, we make sure that any identified radiology concept or word from our negation and uncertainty list is not dropped.

\begin{algorithm2e}[ht]
\DontPrintSemicolon
  \caption{Patient-based Doc-Level CL}
  \KwIn{RadBERT-CL initialized with BlueBert-uncased}
  \KwOut{RadBERT-CL pre-trained using CL}
  \KwData{Preprocessed radiology reports of patients.}
  Initialize the weights of projection head g(.)\\
  \For{each epoch}
  {
  		\While{not converged}{
  		    Sample a mini-batch of training patients $P \in P_{all}$\\
  		    For each $p \in P$, randomly sample two reports ($query$, $key^{+}$) belonging to same patient\\
  		    For each $p \in P$, randomly sample k reports ($key_{-}$) of patients other than $p$\\
  		    Encode $query, key^{+},$ and k-$key^{-}$ with $f(.)$ and $g(.)$\\
  		    Compute loss: $L_{contrastive}$\\
  		    Compute gradient of loss function $\nabla L_{contrastive}$ and update $f(.)$ and $g(.)$\\
  		}
  }
  \textbf{Return} Pre-trained RadBERT-CL

\label{alg1}
\end{algorithm2e}
\subsubsection{Sentence-Level Augmentation}
Sentence-level augmentations are generated by first splitting radiology reports into sentences and then applying random word and phrase dropping  \citep{wu2020clear}, while preserving critical and factual information identified in Info - Preservation module. We propose two different augmentation techniques by associating each sentence with a disease concept from Radlex and a boolean variable indicating presence/absence of any negation or uncertainty phrase. Sentences without any mention of disease concepts are discarded.
\begin{itemize}
    \item \textbf{Disease-based augmentation}: In this technique, we discard all sentences which consist of any mention from our negation or uncertainty dictionary. For a given anchor sentence with disease concept $d$, any other sentence from any report mentioning d can be taken as positive example. Negative samples can be sentences which mention any disease concept except d. Refer Table \ref{tab:augmentationexample} for the example.

    \item \textbf{Disease + Factuality based augmentation}: In this technique, we consider any mention from our negation or uncertainty dictionary along with disease concept while generating augmentation pairs. For a given anchor sentence with disease concept $d$ and negation or uncertainty present, any other sentence from any report mentioning d and negation or uncertainty present can be taken as positive example. Negative samples can be sentences which mention same disease $d$, but negation or uncertainty absent. Refer Table \ref{tab:augmentationexample} for the example.

\end{itemize}

\begin{algorithm2e}[ht]
\DontPrintSemicolon
  
  \KwIn{RadBERT-CL initialized with BlueBert-uncased}
  \KwOut{RadBERT-CL pre-trained using CL}
  \KwData{Preprocessed radiology reports at sentence level: (sentence, disease-mention)}
  Initialize the weights of projection head g(.)\\
  \For{each epoch}
  {
  		\While{not converged}{
  		    Sample a mini-batch of training sentences $S \in S_{all}$\\
  		    For each $s \in S$, randomly sample another sentence ($key^{+}$) with same disease mention\\
  		    For each $s \in S$, randomly sample k sentences ($key_{-}$) having disease mention other than $s$\\
  		    Encode $query, key^{+},$ and k-$key^{-}$ with $f(.)$ and $g(.)$\\
  		    Compute loss: $L_{contrastive}$\\
  		    Compute gradient of loss function $\nabla L_{contrastive}$ and update $f(.)$ and $g(.)$\\
  		}
  }
  \textbf{Return} Pre-trained RadBERT-CL
\caption{Disease-based Sentence-Level CL}
\label{alg2}
\end{algorithm2e}

\begin{algorithm2e}[ht]
\DontPrintSemicolon
  
  \KwIn{RadBERT-CL initialized with BlueBert-uncased}
  \KwOut{RadBERT-CL pre-trained using CL}
  \KwData{Preprocessed radiology reports at sentence level: (sentence, disease-mention, factuality-mention)}
  Initialize the weights of projection head g(.)\\
  \For{each epoch}
  {
  		\While{not converged}{
  		    Sample a mini-batch of training sentences $S \in S_{all}$\\
  		    For each $s \in S$, randomly sample another sentence ($key^{+}$) with same disease and factuality mention\\
  		    For each $s \in S$, randomly sample k sentences ($key_{-}$) having disease and factuality mention other than $s$\\
  		    Encode $query, key^{+},$ and k-$key^{-}$ with $f(.)$ and $g(.)$\\
  		    Compute loss: $L_{contrastive}$\\
  		    Compute gradient of loss function $\nabla L_{contrastive}$ and update $f(.)$ and $g(.)$\\
  		}
  }
  \textbf{Return} Pre-trained RadBERT-CL
\caption{Disease+Factuality-based Sentence-Level CL}
\label{alg3}
\end{algorithm2e}

\vspace{-2em}

\begin{table*}
\centering
\caption{\label{table1}
The weighted F1 scores for fine-tuned RadBERT-CL variants using Model architecture in Figure \ref{fig:structure} (a) and (b). We compare RadBERT-CL variants with SOTA models for reports classification CheXpert \citep{irvin2019chexpert}, and CheXbert \citep{smit2020chexbert} trained on MIMIC-CXR. Reported F1-scores are calculated on the test set of $687$ manually labelled reports, similar to  \citep{smit2020chexbert}. Note that our method does not require any expensive labeled reports during training. Our contrastive pre-training helps RadBERT-CL to outperform the baselines.
}
\begin{tabular}[ht]{llllll}
\hline
\textbf{Category} & \textbf{CheXpert} & \textbf{Previous SOTA} & \textbf{Algorithm 1}& \textbf{Algorithm 2}& \textbf{Algorithm 3}\\
\textbf{} &  & \textbf{CheXbert}  & \textbf{RadBERT-CL}& \textbf{RadBERT-CL}& \textbf{RadBERT-CL}\\
\hline
Enlarged Cardiom. & 0.613 & 0.713 & 0.692 & \textcolor{red}{0.717} & 0.690 \\
Cardiomegaly &  0.764 & 0.815 & 0.808 & 0.806 & \textcolor{red}{0.817} \\
Lung Opacity & \textcolor{red}{0.763}      & 0.741 & 0.761 &  0.747 & 0.746\\
Lung Lesion & 0.683 & 0.664 & \textcolor{red}{0.732} &  0.685 & 0.701\\
Edema & 0.864 & 0.881 & 0.885 &  0.889 & \textcolor{red}{0.891}\\
Consolidation & 0.772 & 0.877 & 0.876 & 0.886 & \textcolor{red}{0.885}\\
Pneumonia & 0.684 & 0.835 & 0.838 &  0.846 & \textcolor{red}{0.847}\\
Atelectasis & 0.917 & \textcolor{red}{0.940} & 0.926 &  0.936 & 0.931\\
Pneumothorax & 0.882 & 0.928 & \textcolor{red}{0.950} &  0.933 & 0.943\\
Pleural Effuison & 0.905 & 0.919 &  0.920 & \textcolor{red}{0.926} & 0.913\\
Pleural Other & 0.478 & 0.534 &   0.541 &  0.577 & \textcolor{red}{0.581}\\
Fracture & 0.671 & 0.791 &  0.791 &  0.796 & \textcolor{red}{0.791}\\
Supported Devices & 0.867 & 0.888 &  0.888 & 0.884 & \textcolor{red}{0.889}\\
No Finding & 0.543 & \textcolor{red}{0.640} & 0.580  &  0.588 & 0.615\\
\hline
\textbf{Average} & 0.743 & 0.798 &  0.799 & 0.801 & 0.804\\
\hline
\end{tabular}
\vspace{0.3cm}

\label{tab:main_result}
\vspace{-0.1cm}
\end{table*}

\subsubsection{Document-Level Augmentation}
Document-level augmentations are generated at the report-level, where each report is first pre-processed with removing extra spaces, newlines, and unwanted tokens. For a given report $r^k$, we apply four types of augmentations (word deletion, span deletion, sentence reordering, and synonym substitution with probability 0.2) mentioned in  \citep{wu2020clear} while preserving critical and factual information identified in Info-Preservation module, to generate positive key. Negative keys can be any report not from the same patient. 

\subsection{Model Architecture}
Our proposed model RadBERT-CL is a two-staged training process: pre-training and fine-tuning (Figure \ref{fig:structure}(a) and (b)). For pre-training, we follow SimCLR  \citep{chen2020simple} framework closely, and use BlueBert architecture as the encoder. Radiology reports are processed by Info-Preservation module and augmentations are generated using techniques proposed in Section \ref{augmentation}. The augmentations are passed through the encoder $f(.)$ and we take the \verb|CLS| output of encoder and further pass it through the projection head $g(.)$. Our projection heads consist of two MLP layers of size 768, along with non-linearity RELU and BatchNorm Layer. After pre-training we discard the projection head and use our pre-trained encoder for fine-tuning.

\subsection{Dataset}

For the disease labelling task, we use MIMIC-CXR dataset  \citep{johnson2019mimiccxrjpg} which consists of $377,110$ chest-Xray images of $227,827$ patients along with their corresponding de-identified radiology reports. The dataset is pseudo-labeled using automatic labeler  \citep{irvin2019chexpert} for the intended set of 14 observations using the entire body of the report.  

In our study, we apply the contrastive pre-training by using the radiology reports from the entire MIMIC-CXR dataset for generating positive and negative augmentations. We divide our dataset into two parts for the fine-tuning stage after removing the duplicate reports of same patient: 80\% for training, 20\% for validation. Note that there is no patient overlap between the training and validation split. Additionally, we have a set of $687$ reports belonging to $687$ unique patients, similar to  \citep{smit2020chexbert}, which has been manually annotated by radiologists for the same $14$ observations, and we evaluate our RadBERT-CL on this dataset. 

\begin{table*}
\centering
\caption{\label{tab:limited}
Transfer learning performance (F1-score) of RadBERT-CL, BERT, and BlueBERT when few labeled data is available. Fine-Tuning is done using randomly selected $400$ reports and F1-score is reported on the remaining $287$ reports of $687$ high-quality manually annotated reports. Reported results are the mean F1-score of the $10$ random training experiments and rounded to 3 decimal places. We identify significant improvements by RadBERT-CL in both Linear Evaluation setting (freeze encoder f(.) parameters and train the classifier layer), and Full-network Evaluation setting (train encoder f(.) and classifier layer end-to-end).
}
\begin{tabular}[ht]{lll}
\hline
\textbf{Model} & \textbf{Linear Evaluation} & \textbf{Full-Network Evaluation}\\
\hline
BERT-\verb|uncased| & 0.137 $\pm 0.012$ & 0.477 $\pm 0.009$ \\
BlueBERT-\verb|uncased| & 0.153 $\pm 0.005$ & 0.480 $\pm 0.007$\\
Algorithm 3 RadBERT-CL & 0.258 $\pm 0.015$ & 0.543 $\pm 0.021$\\
(pre-trained using 687 test reports)\\
Algorithm 3 RadBERT-CL & 0.282 $\pm 0.011$ & 0.591 $\pm 0.019$ \\
(pre-trained using Full MIMIC-CXR unlabelled data)\\
\hline
\end{tabular}
\vspace{0.3cm}

\vspace{-0.5cm}
\end{table*}

\subsection{Contrastive Pre-training}

RadBERT-CL uses a transformer architecture similar to  \citep{peng2019transfer} and pre-trains it using contrastive self-supervised learning similar to \citep{chen2020simple} on MIMIC-CXR dataset. Note that RadBERT-CL can be used on top of other language representation models and is not specific to  \citep{peng2019transfer}. We propose three novel contrastive learning algorithms \ref{alg1},\ref{alg2},\ref{alg3} with the help of augmentation techniques proposed in \ref{augmentation}, which help RadBERT-CL to learn discriminative features across different medical concepts as well as factual cues. As shown in Figure \ref{fig:structure}(a), the augmentation views generated using techniques in \ref{augmentation}, are passed through the our encoder RadBERT-CL $f(.)$ and non-linear projection head $g(.)$ to generate two 768-dimensional vectors $z_i = g(f(view1))$ and $z_j = g(f(view2))$. RadBERT-CL is pre-trained by maximizing the agreement between $z_i$ and $z_j$ using the contrastive loss similar to normalized temperature scaled cross-entropy loss (NT-Xent) \citep{chen2020simple} defined as:
\begin{equation}
L_{(i,j)} = -\log \frac{\exp \left(sim\left(z_i, z_j\right) / \tau\right)}{\sum_{k=1,k\neq i }^{num} \exp \left(sim\left(z_i, z_k\right) / \tau\right)},
\end{equation}
\begin{equation}
    L_{Contrastive} = \sum_{k=1}^{batch-size} L_{(i,j)}
\end{equation}

where $\tau$ is a temperature parameter, and $num$ is the number of negative views. We calculate the loss for each sample in our mini-batch and sum them to estimate $L_{Contrastive}$. We calculate the gradient $\nabla L_{contrastive}$ and back-propagate it to update our encoder $f(.)$ and $g(.)$. Contrastive learning benefits from training for larger epochs  \citep{he2020momentum, chen2020simple, grill2020bootstrap}, so we trained RadBERT-CL for 100 epochs using SGD optimizer.
Note that after pre-training, we discard the project head $g(.)$ and only use our encoder $f(.)$ for fine-tuing on downstream task.

\subsection{Supervised Fine-Tuning}
In order to use the pre-trained RadBERT-CL model for our downstream task of report classification, we further fine-tune $f(.)$ on the pseudo-labels of radiology report classification task as shown in Figure \ref{fig:structure}(b). Our disease is multi-class classification problem and We use cross-entropy loss as our supervised classification loss, defined as:
\begin{equation}
L^i_{l,k} = \sum_{l}\sum_{k} y^i_{l,k} \times log(\tilde y^i_{l,k})
\end{equation}
\begin{equation}
    L_{classification} = \sum_{i=1}^{batch-size} L^i_{l,k}
\end{equation}
where, $i$ denotes $i-th$ training example, $l$ denotes class label (Edema, Cardiomegaly, etc.), $k$ $\in$ \{Positive, Negative, Uncertain, Blank\}. We calculate the gradient $\nabla L_{classification}$ and back-propagate it to update our encoder $f(.)$.

\vspace{-1em}

\section{Evaluation and Results}

\subsection{Evaluation}
Following  \citep{smit2020chexbert}, we evaluate our system based on its average performance on three retrieval tasks: positive extraction, negative extraction, and uncertainty extraction. For each of the $14$ observations, we compute a weighted average of the F1 scores on each of the above three tasks, weighted by the support for each class of interest, which we call the weighted-F1 metric. Table \ref{tab:main_result} presents the weighted-F1 score of RadBERT-CL using our three different variants of contrastive learning and their comparisons with SOTA methods. We have also presented the detailed evaluation score of our best RadBERT-CL variant (Algorithm \ref{alg3}) for all three retrieval tasks in Appendix Table \ref{tab:detailed_result}.

\begin{table*}
\caption{Cosine Similarity between the normalized-[CLS] embeddings of report snippets generated by RadBERT-CL after contrastive pre-training. Our RadBERT-CL embeddings are capable of distinguishing between the factual nuances of medical reports which cannot be captured by the embeddings generated by BERT, and BlueBert. Our model is able to capture fine-grained differences among diseases, negation, and uncertainty in the latent representations.}
\centering
\begin{tabular}{llll}
\hline
\textbf{Report Segment} & \textbf{BERT} & \textbf{BlueBert}& \textbf{Algorithm \ref{alg3}}\\
 & & &\textbf{RadBERT-CL}\\
\hline
\textit{... definite focal consolidation \textcolor{red}{is seen} in left side of lungs...} & 0.9411 & 0.9223 & -0.8266 \\
\textit{... the lungs are \textcolor{red}{clear of} any focal consolidation ...} & & \\
\hline
\textit{... subtle opacity at the right base \textcolor{red}{could represent} infection ...} & 0.9120 & 0.9038 & 0.4332\\
\textit{... patchy left base opacity \textcolor{red}{represent} severe infection ... }\\
\hline
\textit{... pleural \textcolor{red}{effusion} is obserevd ...} & 0.9752 & 0.8931 & 0.3836\\
\textit{... pleural \textcolor{red}{edema} is seen ...} & &\\
\hline
\end{tabular}
\vspace{-0.3cm}

\label{tab:embedding}
\end{table*}

To demonstrate the effectiveness of RadBERT-CL performance when only a few labeled data is available, we evaluated RadBERT-CL performance in two different training scenarios: (a) pre-train RadBERT-CL using Algorithm 3 on 687 high-quality annotated dataset (no manually annotated label is used), fine-tune on randomly selected 400 high-quality annotated dataset, and test it on remaining 287 high-quality annotated dataset. (b) pre-train RadBERT-CL using Algorithm 3 on entire MIMIC CXR, fine-tune on randomly selected 400 high-quality annotated dataset, and test it on remaining 287 high-quality annotated dataset.

\vspace{-1em}

\subsection{Results}
We observe that our RadBERT-CL model pre-trained using Algorithm \ref{alg3} outperforms previous state-of-the-art model CheXbert in $7$ out of $14$ findings after fine-tuning. Table \ref{tab:main_result} presents the weighted F1 scores of RadBERT-CL varients and previous SOTA systems CheXpert and CheXbert. Our model variants combined together outperform CheXbert in $11$ out of $14$ findings. Note that CheXbert training is calibrated under the supervision of \textbf{$\sim 1000$ manually annotated reports} by radiologists while our system is trained using weakly labeled reports. With the help of the guided-supervision of expert-level annotated data as proposed in CheXbert \citep{smit2020chexbert}, we believe that our system will show more significant improvements.

In our analysis using Algorithm \ref{alg1},\ref{alg2},\ref{alg3}, we found that RadBERT-CL is very successful in capturing the factual information present in radiology reports. We calculated the cosine similarity between \verb|CLS| embeddings generated by two factually different report snippets as shown in Table \ref{tab:embedding}, by BERT, BlueBert and RadBERT-CL. RadBERT-CL is able to distinguish between the factual nuances of medical reports which are not captured in the representations generated by BERT and BlueBert.

While deep learning methods often require expert-annotated high-quality data for training, getting sufficiently annotated data in the medical domain is very costly due to the limited availability of human experts. However, we have enough unlabelled data which can be used to improve our DL models with the supervision of few high-quality annotated data. Table \ref{tab:limited} illustrates our RadBERT-CL performance in such scenario. Clearly, our model outperforms conventional fine-tuning using BERT/BlueBert for the classification task, by huge margins of $0.06$ to $0.11$ on weighted F1-metric. Better performance in Linear evaluation settings indicates that the representations learned by RadBERT-CL in pre-training stage are significantly better than BERT/BlueBert. Our experiments confirm that using largely available unsupervised data to pre-train transformers using contrastive learning provide significant improvement in fine-tuning tasks when few labelled data is available. 

\vspace{-1em}

\section{Conclusion}
In this work, we present novel data augmentation techniques for contrastive learning to capture factual nuances of medical domain. Our method involves pre-training transformers using abundance of unsupervised data to capture fine-grained domain knowledge before fine-tuning it for downstream tasks such as disease classification. We further show that such training strategy improves the performance in downstream tasks significantly in limited data settings.  We hope that this work can draw community attention towards the ability of contrastive learning to capture discriminative properties in the medical domain.

\bibliography{references}

\section{Appendix}

\begin{table}[h]
\vspace{-1em}
\centering
\begin{tabular}{lll}
\hline
\textbf{Hyperparameter} & \textbf{Pretraining}& \textbf{Finetuning}\\
\hline
batch-size & 128 & 32 \\
learning-rate & 0.1 & 2e-5 \\
optimizer & SGD & Adam \\ 
temperature (CL) & 0.4 & - \\ 
n\_epochs & 100 & 10 \\
beta & - & [0.9, 0.99]  \\ 
Aug. Probability & 0.2 & -  \\ 
\hline

\end{tabular}
\caption{Training details for RadBERT-CL Pretraining and Finetuning Stages.}
\label{tab:train_params}
\end{table}

\begin{figure*}[h]
\includegraphics[width=\textwidth]{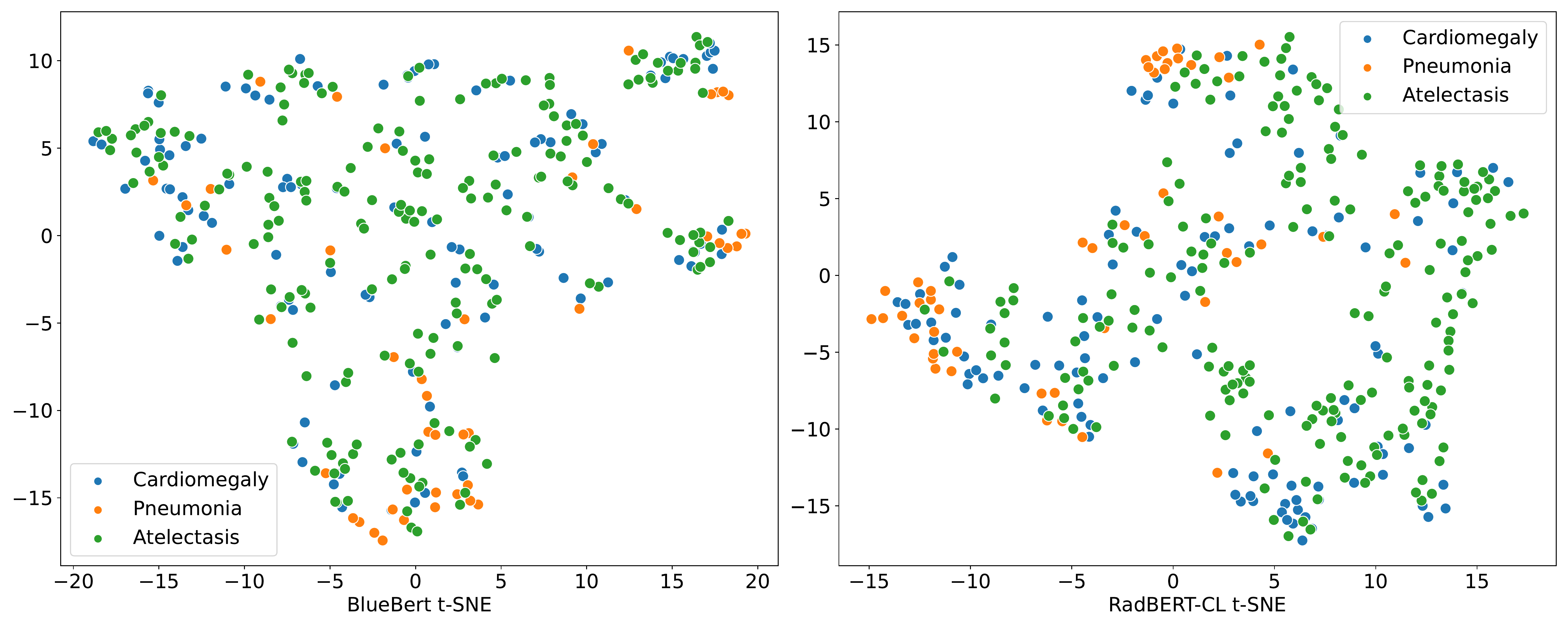}
\caption{t-SNE visualization of BlueBert and RadBERT-CL(Algorithm \ref{alg3}) for radiology reports annotated positive for three major diseases (Cardiomegaly, Pneumonia, and Atelectasis). Note that the reports used for generating the t-SNE plot are sampled from 687 radiologists annotated test set which are not used in RadBERT-CL pre-training. From the figure, it is evident that embeddings generated after pre-training RadBERT-CL with contrastive learning, is more informative compared to BlueBert on unseen data. }
\label{fig:tsne}
\end{figure*}

\begin{table*}[h]

    \begin{tabular}{p{\textwidth}} 
    \hline
       
        \textbf{Report Snippet:} \textit{... apparent new small right pleural edema manifested by posterior blunting of right costophrenic sulcus ...}\\
        \vspace{0.001cm}
        \textbf{Prediction:} \textcolor{red}{Pleural Other}\\
        \textbf{Ground Truth:} \textcolor{blue}{Edema} \\
        \textbf{Reasoning:} the presence of \textit{pleural} keyword along with edema may have confused the model to classify it as Pleural Other.\\
        \vspace{0.001cm}
        
        \vspace{0.1cm}
        \textbf{Report Snippet:} \textit{... new area of pleural abnormality has developed in right side of lungs, and the heart and mediastinal structures and bony structures remain normal in appearance ...}\\
        \vspace{0.001cm}
        \textbf{Prediction:} \textcolor{red}{Pleural Effusion}\\
        \textbf{Ground Truth:} \textcolor{blue}{Pleural Other} \\
        \textbf{Reasoning:} we found in reports that many pleural disorders share similar context which possibly make it difficult to classify them correctly. This can also explain the low F1-score of Pleural Other category.\\
        \vspace{0.001cm}
        
        \vspace{0.1cm}
        \textbf{Report Snippet:} \textit{... mild interstitial edema and small right pleural effusion are new since \_\_\_ ...}\\
        \vspace{0.001cm}
        \textbf{Prediction:} \textcolor{blue}{Pleural Effusion}\\
        \textbf{Ground Truth:} \textcolor{blue}{Pleural Effusion, Edema} \\
        \textbf{Reasoning:} the model misses to identify edema and only identified Pleural Effusion possibly because majority of times, edema is mentioned as Pleural Edema in reports.\\
        \vspace{0.001cm}
   
    \end{tabular}
    
    \caption{Examples where RadBERT-CL incorrectly assign or misses label while making prediction. We include speculative reasoning for the classification errors.}
    \label{tab:reasoning}
   
\end{table*}

\begin{table*}
\centering
\begin{tabular}[ht]{lllll}
\hline
\textbf{Category} & \textbf{Positive F1} & \textbf{Negation F1} & \textbf{Uncertain F1}& \textbf{Blank F1}\\

\hline
Enlarged Cardiomediastinum & 0.579 & 0.786 & 0.831 & 0.965 \\
Cardiomegaly &  0.870 & 0.862 & 0.433 & 0.978\\
Lung Opacity &  0.820 & 0.200 &  0.512 & 0.910\\
Lung Lesion & 0.777 & 0.571 &  0.211 & 0.983\\
Edema & 0.913 & 0.901 & 0.745 &  0.993 \\
Consolidation & 0.909 & 0.824 & 0.876 & 0.997 \\
Pneumonia & 0.786 & 0.916 & 0.807 &  0.991 \\
Atelectasis & 0.962 & 0.444 &  0.874 & 0.999\\
Pneumothorax & 0.850 & 0.971  &  0.526 & 0.996\\
Pleural Effuison & 0.938 & 0.957 &  0.596 & 0.985\\
Pleural Other & 0.623 & 0.234 &   0.114 &  0.981 \\
Fracture & 0.894 & 0.333 &  0.667 &  0.993 \\
Supported Devices & 0.902 & 0.100 &  0.000 & 0.942 \\
No Finding & 0.592   & 0.000  &  0.000 & 0.978\\
\hline
\end{tabular}
\vspace{0.3cm}
\caption{\label{table1}
Detailed F1-evaluation of RadBERT-CL variant (Algorithm \ref{alg3}) for the classification tasks of positive extraction, negation extraction, uncertainty extraction, and blank for each of our 14 observations. Note that for "Blank", we have f1-scores related to positive extraction and blank, while the other two are set to zero.  
}
\label{tab:detailed_result}
\vspace{-0.1cm}
\end{table*}

\begin{table*}
    \centering
    \caption{Examples from the set of rules in our Info-Preservation Module for Negation and Uncertainty Detection and their corresponding matching sentences.}
    \begin{tabular}[ht]{l}
    \toprule
    \textbf{a. Negation Detection}  \\
    \midrule
     
    \textbf{RULE:} $^* +\ clear/free/disappearance\ +\ <prep\_of>\ +\ ^* +\ DISEASE\_CONCEPT$ \\
    1. the left lung is \underline{free of} consolidations or pneumothorax \\
    2. the lungs are \underline{clear of} any focal consolidation\\\vspace{0.2cm}
    3. pleural sinuses are \underline{free of} any fluid accumulation\\
    
    \textbf{RULE:} $^* +\ no/not\ +\ evidence/*\ +\ <prep\_[of|for]>\ +\ ^* +\ DISEASE\_CONCEPT$ \\
    1. within the remaining well-ventilated lung, there is \underline{no evidence of} pneumonia \\
    2. there is \underline{not evidence for} pulmonary edema\\
    3. there are \underline{no evidences of} acute pneumothorax\\
    
    \midrule
    \textbf{b. Uncertainty Detection}  \\
    \midrule
    
    \textbf{RULE:} $^* +\ could be/may be/...\ +\ ^* +\ DISEASE\_CONCEPT$\\
    1. there are bibasilar opacities which \underline{could be} due to atelectasis given low lung volumes \\
    2. perihilar opacity \underline{could be} due to asymmetrical edema\\
    3. left base opacity \underline{may be} due to atelectasis\\
    
    \textbf{RULE:} $^* +\ suggest/suspect/[-ing|-ed]\ +\ ^* +\ DISEASE\_CONCEPT$\\
    1. signs of parenchymal changes \underline{suggesting} pneumonia \\
    2.  the left heart border is silhouetted, with a \underline{suspected} left basilar opacity\\
    3. prominence of the central pulmonary vasculature \underline{suggesting} mild pulmonary edema \\
    \hline
    \end{tabular}
    \vspace{0.3cm}
    
    \label{tab:ruleexample}
\end{table*}

\end{document}